%% file: root.tex
\documentclass[letterpaper, 10 pt, conference]{ieeeconf} 
\usepackage{times}

\usepackage{multicol}
\usepackage{graphicx}
\usepackage{wrapfig}
\usepackage[font={small}]{caption}
\usepackage{cite}
\usepackage{subcaption}
\usepackage{amsmath}
\usepackage{bbm}
\usepackage{amssymb}
\usepackage{gensymb}
\usepackage{dblfloatfix} 
\usepackage{dsfont}
\usepackage{multirow}
\usepackage{hyperref}
\usepackage{xcolor}
\IEEEoverridecommandlockouts                              

\title{\LARGE \bf OGMP: Oracle Guided Multi-mode Policies for \\Agile and Versatile Robot Control}

\author{
Lokesh Krishna, Nikhil Sobanbabu, and Quan Nguyen 
\thanks{All authors with the Dynamic Robotics and Control  Laboratory, University of Southern California, Los Angeles, CA 90089, USA 
{\tt\small lkrajan@usc.edu, ns\_562@usc.edu, quann@usc.edu}}
}

\begin{document}
\maketitle
\begin{abstract}

The efficacy of reinforcement learning for robot control relies on the tailored integration of task-specific priors and heuristics for effective exploration, which challenges their straightforward application to complex tasks and necessitates a unified approach. In this work, we define a general class for priors called oracles that generate state references when queried in a closed-loop manner during training. By bounding the permissible state around the oracle's ansatz, we propose a task-agnostic oracle-guided policy optimization. To enhance modularity, we introduce task-vital modes, showing that a policy mastering a compact set of modes and transitions can handle infinite-horizon tasks. For instance, to perform parkour on an infinitely long track, the policy must learn to jump, leap, pace, and transition between these modes effectively. We validate this approach in challenging bipedal control tasks: parkour and diving—using a 16-DoF dynamic bipedal robot, HECTOR. Our method results in a single policy per task, solving parkour across diverse tracks and omnidirectional diving from varied heights up to $2m$ in simulation, showcasing versatile agility. We demonstrate successful sim-to-real transfer of parkour, including leaping over gaps up to $105\%$ of the leg length, jumping over blocks up to $20\%$ of the robot's nominal height, and pacing at speeds of up to $0.6$ m/s, along with effective transitions between these modes in the real robot. 


\end{abstract}

\input{sections/introduction}

\input{sections/proposed_framework}
\input{sections/method}

\input{sections/results}
\input{sections/conclusion}

\input{sections/appendix}
 \bibliographystyle{ieeetr}
\bibliography{references}
\end{document}

%% file: sections/introduction.tex
\section{Introduction}

Deep reinforcement learning (RL) has shown remarkable success in synthesizing control policies for hybrid and underactuated legged robots \cite{siekmann2021blind}, particularly in enabling inherently stable quadrupedal robots to achieve extreme parkour \cite{hoeller2024anymal,cheng2023extreme, zhuang2023robot}, agile \cite{rudin2022advanced}, and robust \cite{miki2022learning} locomotion. Following a common philosophy: 1) define an exhaustive observation space, 2) engineer task-specific rewards and/or curriculum, 3) perform policy distillation, and 4) extensively randomize, these methods rely on task-specific tricks in each step,  lacking a systematic approach for robot control. 
Specifically, since Deep RL methods are quasi-solvers for unconstrained optimization, they are prone to anomalous, case-specific local optima. 
Hence, practitioners often resort to task-specific reward shaping \cite{hoeller2024anymal,rudin2022advanced,miki2022learning} and heuristics \cite{cheng2023extreme} for a structured exploration and to meet the intended performance. Furthermore, the established approach for robot control is privileged training and policy distillation: training teacher policies \cite{cheng2023extreme,zhuang2023robot} with privileged information, solving a pseudo-MDP with RL, and distilling it into a single policy for the true POMDP. In contrast, we aim to find an optimal policy by structured exploration in the true POMDP  guided by priors. 

\begin{figure}
	\includegraphics[width=0.48\textwidth]{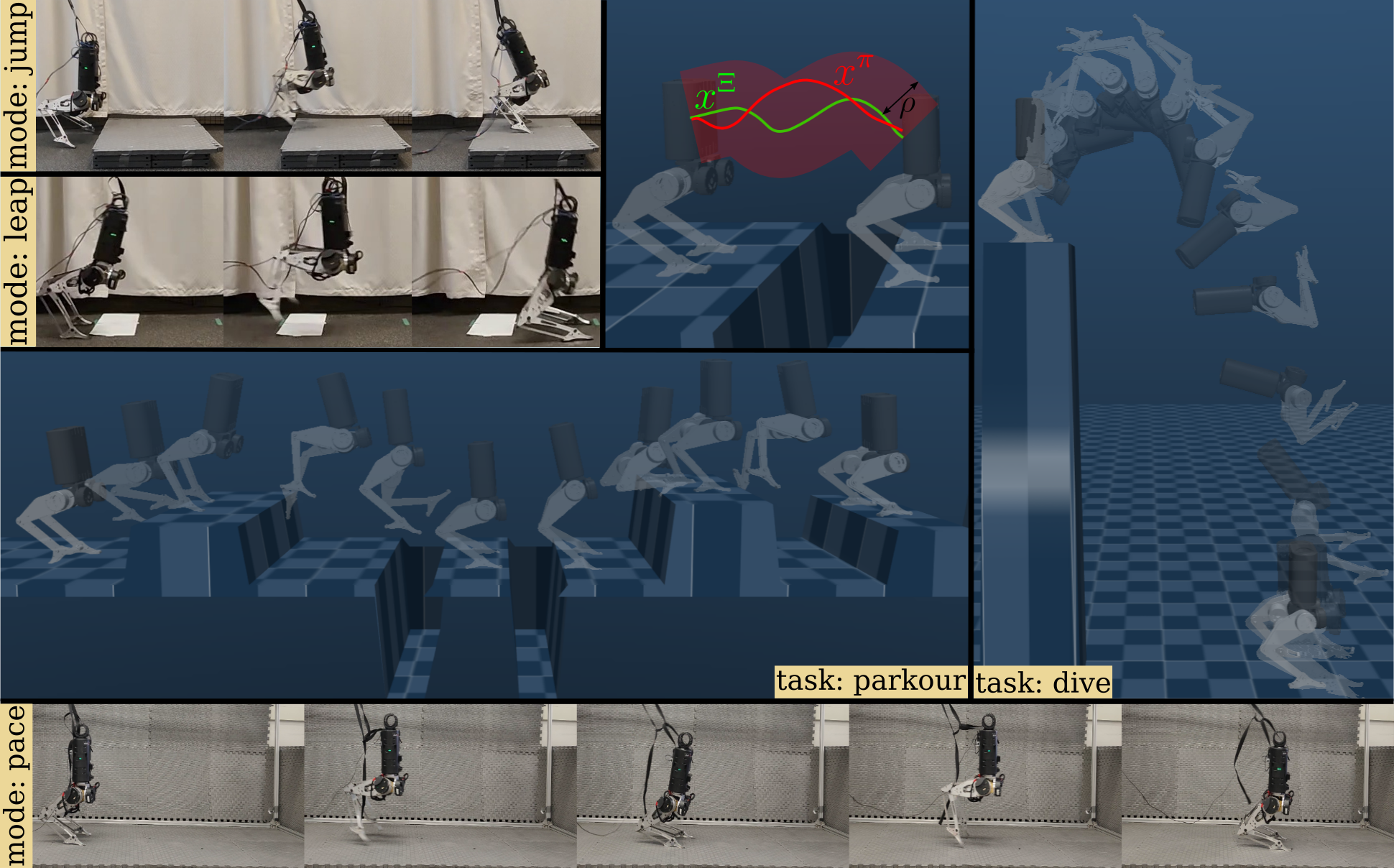}
	\caption{ Overview of OGMP: Oracle guided policy optimization and the applied tasks visualized. Trained OGMPs $\pi_\text{parkour}$ performing agile parkour in the simulation and the real-robot. $\pi_\text{dive}$ performing a frontflip dive from a $2$m high platform in simulation. Accompanying video results at : $\href{https://youtu.be/69SVc-43Oqg?si=w4r3i67oBaoThLN7}{link}$
 }
	\label{fig:ov}
	\vspace{-6mm}
\end{figure}

Guided Policy Optimization (GPO) aims to improve sample efficiency and mitigate poor local optima. GPO methods are either \emph{Control-guided} or \emph{Reference-guided}. \emph{Control-guided} approaches require control trajectories: \cite{levine2013guided, levine2014lerning} employ prior controllers and policy-trajectory constraints, while \cite{mordatch2014combining, cheng19control} alternate trajectory optimization and minimize RL variance. \cite{carius2020mpcnet, gangapurwala2020guided, jenelten2023dtc} demonstrate quadrupedal locomotion but relying on pre-existing model-based controllers, limiting complex tasks (e.g., parkour). \emph{Reference-guided} methods like \cite{peng2018deep, peng2021amp, vollenweider2022advanced, li2024reinforcement, luo2024smpl} use morphologically similar state-reference trajectories to guide RL to learn the corresponding optimal actions for character control. However, with pre-generated open-loop reference, policy exploration is confined to the demonstration's scenario, thus hindering the emergent behaviors(like recovery) seen in from-scratch RL methods \cite{siekmann2021blind, cheng2023extreme, hoeller2024anymal}, which explore full-order dynamics and challenging randomization in simulation, crucial for real robot control. Therefore, we propose a reference-guided policy optimization using closed-loop state-reference generators(oracles) that can be queried dynamically to produce references from any state, with a novel hyperparameter to address local optima in complex tasks.


 Alternatively, imitation learning (IL) proves to be a reliable task-agnostic strategy for robot manipulation with dynamically consistent demonstrations through proficient human teleoperation, which solves the intended task~\cite {chi2023diffusion}. In contrast, we have dynamically inconsistent demonstrations for locomotion that partially solve a task, challenging the direct application of IL. For instance, to parkour with a bipedal robot, we may have demonstrations for runs, leaps, and jumps from motion capture on humans, which suffer from morphological dissimilarity due to source-target mismatches  \cite{bohez2022imitate}. Moreover, naive imitation of partial demonstrations (run, leap, etc.) does not guarantee to solve the overarching task (parkour), requiring high-level RL-trained policy for transitions and emergent behaviors \cite{haarnoja2024learning,bohez2022imitate, hasenclever2020comic}. Besides demonstrations, robot tasks have rich priors like heuristics \cite{raibert1986legged}, task/motion planners \cite{norby2020fast}, and model-based controllers\cite{f&mmpc}, which can guide learning, leading to regularized behaviors \cite{jenelten2023dtc}. While the idea of imitating such priors has been studied,  we instead propose building a ``trust region" in the state space around the prior's solution.  Thus, the more we ``trust'' a prior, the tighter the trust region could be and vice versa. Formally defining a general class for priors: oracle, an oracle-guided policy optimization can be performed by bounding the policy's permissible state space within the local neighborhood of an oracle's ansatz. Empirically, we observe that the right choice of this bound helps escape erroneous local optima providing an optimal balance between emergent and regularized behaviors ideal for robot control, making it an effective hyperparameter in practice.

On the other hand, solving complex tasks requires behavioral multi-modality. Classical multi-mode control \cite{koo2001multi,asarin2000effective}involves switching among a finite set of pre-designed controllers to address high-level tasks, creating a pseudo-hybrid system. Learning methods leave multi-modality in control to emerge implicitly \cite{rudin2022advanced, cheng2023extreme}, lacking a methodical synthesis. \cite{hasenclever2020comic} propose encoding a dataset of demonstrations to a latent space and latent conditioning to train multi-skilled policies. However, the notion of skill is poorly defined. For instance, solving a task requires not only mastering discrete modes (e.g., walking, jumping) but also continuous parameter variation of the same (e.g., speed, height) and inter-skill transitions. \cite{hoeller2024anymal} trains multiple low-level controllers and a high-level mode-switching policy for quadrupedal parkour, requiring diverse reward shaping and training routines. \cite{krishna2023learning} shows that a ``fixed" set of uni-mode controllers limits complex transition maneuvers, introducing a single multi-mode policy for mastering a set of modes and transitions to handle new tasks zero-shot. In line with this approach, we aim to achieve a single policy that learns a finite set of modes with ``infinite" parameter variations and transitions through reference-guided policy optimization. Unlike switching fixed controllers, we hypothesize that reference-guided exploration can better accommodate emergent control modes. To this end, we first show task-vital multi-modality as a way to decompose tasks into their principal modes and transcribe them into our proposed OGMP framework. Thus, the major contributions of our paper are twofold: 
\begin{itemize}
  
    \item Oracle Guided Multi-mode Policies: A theoretical framework for task-centered control synthesis leveraging oracle-guided optimization to effectively search through bounded exploration and task-vital multi-modality for versatile control.
    
    \item Experimental validation on agile bipedal control tasks requiring versatility, such as parkour and diving. A single policy per task demonstrates the ability to perform diverse variants of the task-vital modes realized in simulation and the real-robot on the 16-Dof bipedal robot Hector. 
    
\end{itemize}

The remaining paper is structured as follows: Sec. \ref{sec:ogmp} presents the theoretical development of our framework, Sec. \ref{sec:method} discusses applying the proposed framework to bipedal control tasks, followed by Sec. \ref{sec:results} presenting the experimental results, analysis and ablation studies.

%% file: sections/proposed_framework.tex
\section{Oracle Guided Multi-mode Policies}
\label{sec:ogmp}

This section presents our theoretical framework with two synergetic ideas: oracle-guided policy optimization and task-vital multi-modality.
Specifically, we aim to prune undesirable local optima by bounding exploration to the local neighborhood of an oracle and by designing the learning of multiple behavior modes and transitions to effectively solve tasks.

\subsubsection{Oracle Guided Policy Optimization(OGPO)} Let $\mathcal{T}$ be an infinite horizon task with a task parameter set, $\psi_{\mathcal{T}} \in \Psi_{\mathcal{T}}$. Sufficiently solving $\mathcal{T}$ requires maximizing a task objective, $J_\mathcal{T}$ over the task parameter distribution, $ p (\psi_{\mathcal{T}})$. Given the corresponding state space (or task space) of interest, $x \in \mathcal{X}$ let  $x_t$,\, $x[a,b]$ denote a state at time $t$ and a state trajectory from time $t\in [a,b]$ respectively. We define $\Xi$ to be a receding horizon oracle that provides a finite horizon state trajectory ($x^{\Xi}[t,\,t+\Delta t]$) from any given state ($x_t$) until $\Delta t$ into the future for any task-variant ($\psi_{\mathcal{T}}$), such that 
$x^{\Xi}[t,\,t+\Delta t]$ is always within an $\epsilon\text{-neighborhood}$ of an optimal state trajectory. Formally,
\begin{subequations}
\begin{align}
 \quad & x^{\Xi}[t,\,t+\Delta t] = \Xi(x_t, \psi_{\mathcal{T}})\\
 \textrm{s.t.} 
 \quad &\exists \,\, x^{\Xi}[t,\,t+\Delta t] 
 \quad \forall \, (x_t, \psi_{\mathcal{T}}) \\
 \quad &\|x^{\Xi}_t - x^*_{t}\|_W<\epsilon \,\,\,\,\, \forall \, t\in [0,\,\infty)
\end{align}
\label{eq:oracle_def}
\end{subequations}
where $\epsilon$ is the maximum deviation bound, a constant for a given pair $(\mathcal{T},\,\Xi)$, and $W$ is a diagonal weight matrix\footnote{Note that $x^*[0, \infty],\, \epsilon$ are unknown and are only provided for constructing the conceptual argument}. We aim to obtain the optimal policy $\pi^*$  from a policy class $\Pi$ guided by $\Xi$ that sufficiently solves $\mathcal{T}$. 
Since $\Xi$ provides a reference in the state space, we propose constraining the permissible states for $\pi$ to be within a $\rho$-neighbourhood of the oracle's guidance. Formally, 
\begin{subequations}
\vspace{-1mm}
\begin{align}
\quad & \pi^* := \arg \max_{\pi \in \Pi} J_\mathcal{T} \\
\textrm{s.t.}
\quad & \|x^{\pi}_t - x^{\Xi}_t\|_W< \rho \quad \forall t \in [0,\infty)
\end{align}
\end{subequations}
Where $x^{\pi}$ are the states visited while rolling out policy $\pi$, $\rho$ is the permissible state-bound for oracle-guided exploration. 
\begin{figure}
    \centering
    \begin{subfigure}[t]{0.2\textwidth}
        \centering
        \includegraphics[width=0.8\textwidth]{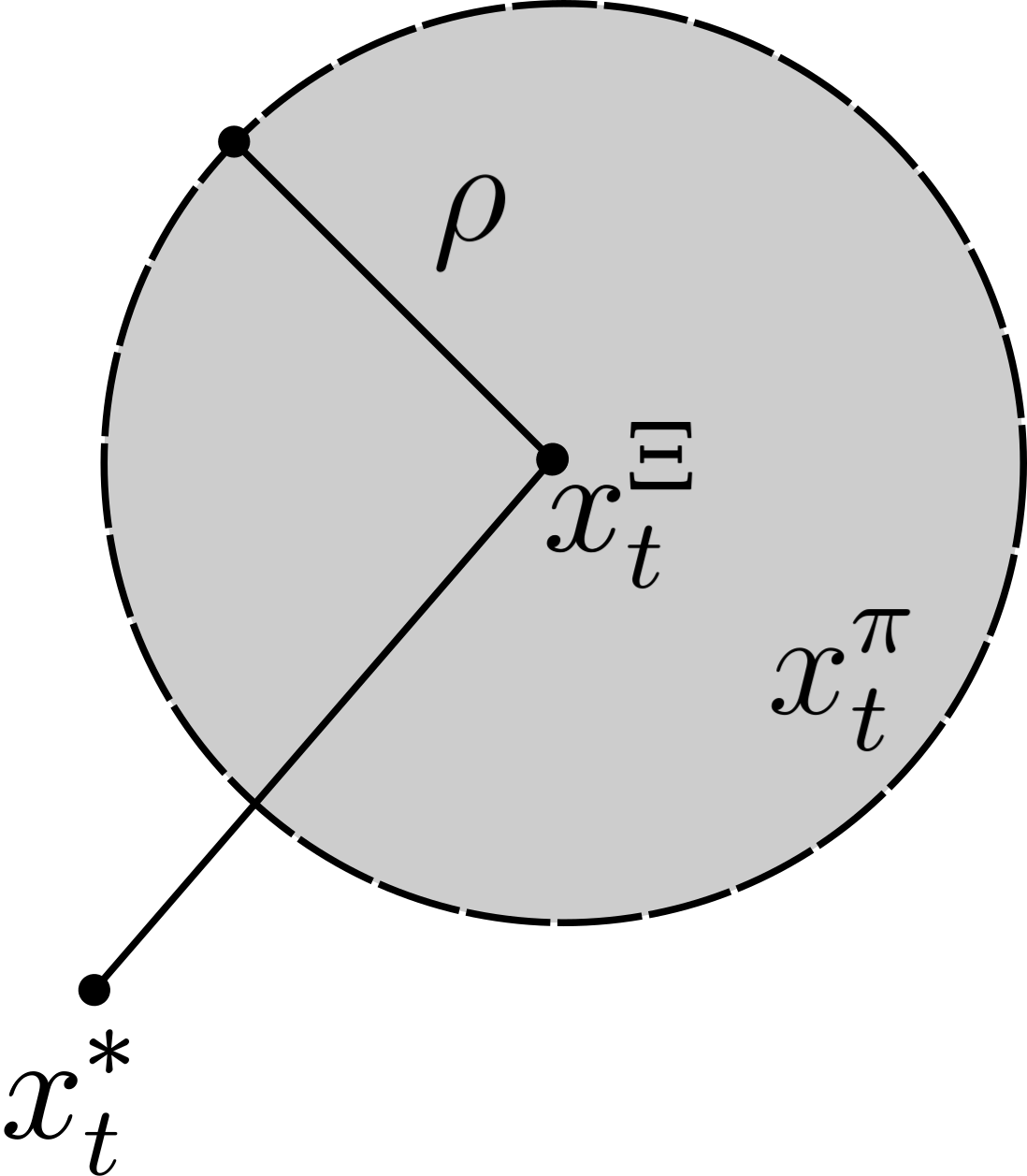}
        \caption{permissible states of $\pi$ within $\rho$-neighbourhood of $x^\Xi_t$ }
        \label{fig:perm_state}
    \end{subfigure}
    \begin{subfigure}[t]{0.24\textwidth}
        \vspace{-3cm}
        \centering
        \includegraphics[width=\textwidth]{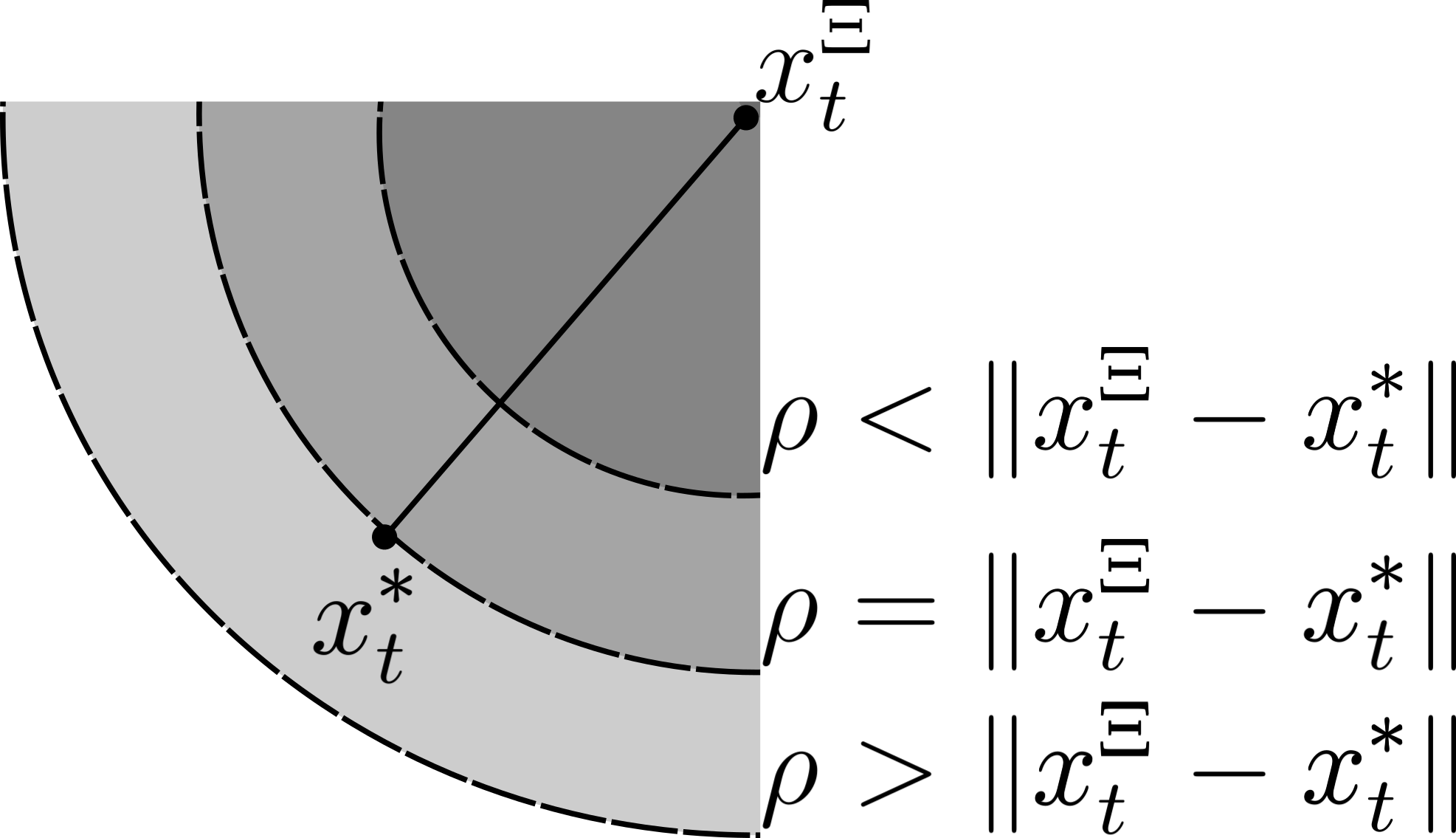}
        \caption{Different choices of $\rho$ }
        \label{fig:diff_rho}
    \end{subfigure}
    \caption{Oracle bounded exploration}
    \label{fig:obe}
\vspace{-6mm}
\end{figure}
With the above setup, one can observe that the bounded set of permissible states visualized in Fig. \ref{fig:perm_state} is given by 
\begin{equation}
        x^{\pi}_t \in \{x \, | \, \| x^{\Xi}_t - x^*_t \| - \rho \leq \| x - x^*_t \| \leq \| x^{\Xi}_t - x^*_t \| + \rho \}
\label{eq:oracle_neigbourhood}
\end{equation}

Thus, $\rho$ should be chosen so that $x^*_t$ is reachable by $x^{\pi}_t$, which requires the lower bound in Eq. \ref{eq:oracle_neigbourhood}, to be non-positive. Therefore as shown in Fig \ref{fig:diff_rho}, for  $x^*_t$ to be within the permissible states of $\pi$, $\rho$ must satisfy
\begin{equation}
        ( x^*_t \text{ is reachable by } x^{\pi}_t ) \implies \rho \geq  \| x^{\Xi}_t - x^*_t \| 
        \label{eq:reachable}
\end{equation}
By definition (Eq. \ref{eq:oracle_def}.c), since $\| x^{\Xi}_t - x^*_t \| < \epsilon$ a sufficient choice of $\rho$ to satisfy Eq \ref{eq:reachable} for all time , is 
\begin{equation}
    \rho \geq \epsilon
\label{eq:rho_suff}
\end{equation}
$\epsilon$ being maximum deviation of $x^\Xi$, an oracle with low $\epsilon$ will generate references closer to $x^*$ thus, the exploration can be bounded to tight-neighborhood to filter out most local optima in the objective landscape
In contrast, for a ``poor" oracle with high $\epsilon$, there should be sufficient search space for $\pi$ to explore around $x^\Xi$ and converge to $x^*$. 
From Eq \ref{eq:rho_suff}, as $\epsilon \to \infty$, the optimization is unguided, thus needing $ \rho \to \infty$ (a standard RL setting). 
Conversely, as $\epsilon \to 0$, an arbitrarily small $\rho$ satisfying Eq \ref{eq:rho_suff} could be chosen to avoid local optima by localizing the search while still being able to converge  $x^*_t$.
In practice, as $\epsilon$ is unknown, we perform a grid search over $\rho$ for any given $(\mathcal{T},\,\Xi)$ .

\subsubsection{Task Vital Multi-modality}
A policy learning to solve a task, $\mathcal{T}$, can be seen as mastering a ``bundle" of spacetime trajectories in the task space, corresponding to $\psi_{\mathcal{T}} \in \Psi_{\mathcal{T}}$. Simple tasks allow straightforward oracle construction satisfying Eq \ref{eq:oracle_def}. However, complex and infinite-horizon tasks make $\Psi_{\mathcal{T}}$ intractable. For instance, an oracle for indefinite parkour requires knowledge of an infinite track apriori, which is impractical. To address this, we define modes as finite spacetime segments that preserve some spatial and/or temporal invariances. Like in parkour, modes like jump and leap remain consistent regardless of location or time. Therefore, we define a finite set of modes, $\mathbb{M}$, having a temporal length of $\Delta t$, vital for $\mathcal{T}$. Each $m \in \mathbb{M}$ (like jump) can have continuous parameters $\Psi_{m}$ (like jump height). Then the mode parameter set $\Psi_{\mathbb{M}} := \bigcup_{i=1}^{|\mathbb{M}|} \Psi_{m_i}$ is related to the task parameter set as $\Psi_{\mathcal{T}} := \Psi_{\mathbb{M}}^{N}$, where $N$ is the number of horizons. Thus, by mastering modes in $\mathbb{M}$ (jump, leap, and pace) and transitions over varying $\Psi_{\mathbb{M}}$ (speeds, distances, and heights ), task $\mathcal{T}$ (indefinite parkour, $N \rightarrow \infty$) can be solved as visualized in Fig \ref{fig:afo}.a.

%% file: sections/method.tex
\section{Design Methodology}
\label{sec:method}

\begin{figure*}[t!]
	\centering
	\includegraphics[width=\textwidth,height =0.4\textwidth]{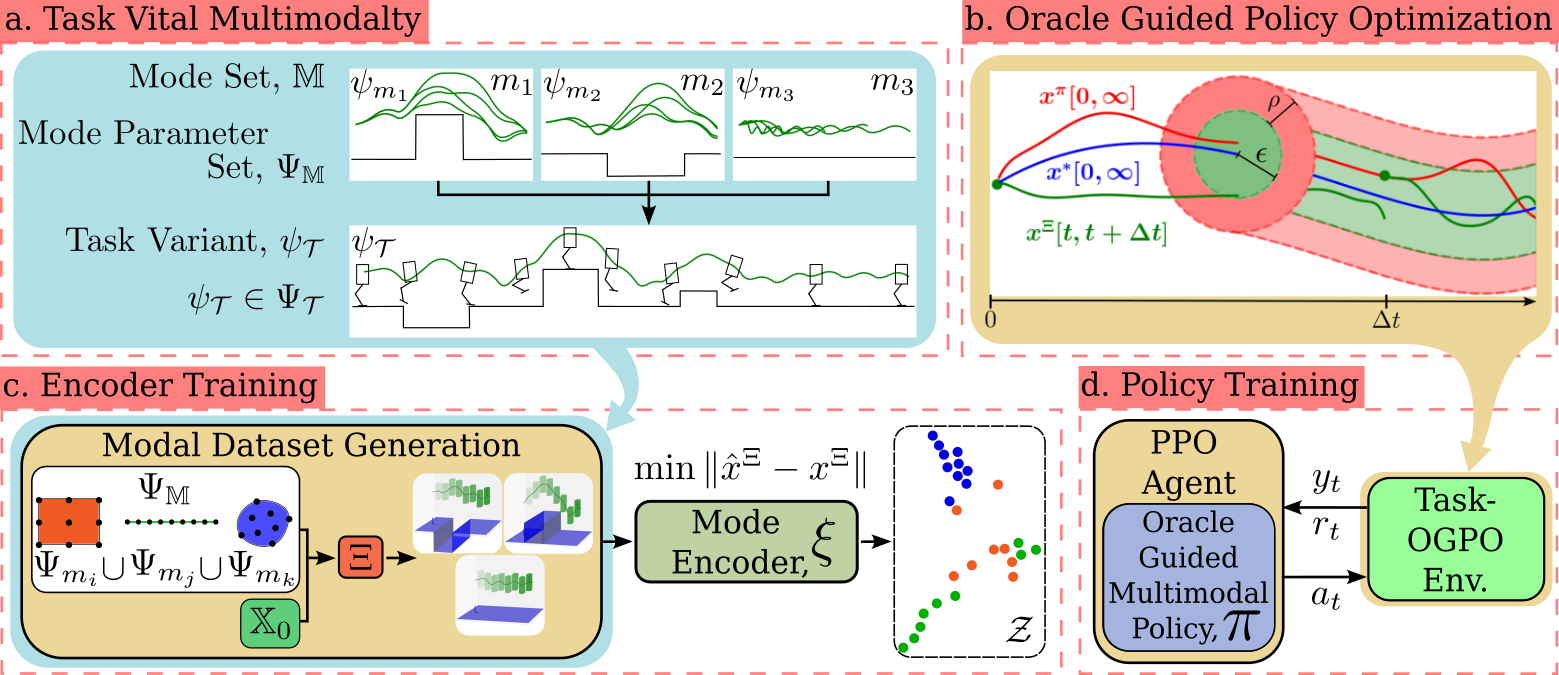}
	\caption{Overview of the design methodology: a) The breakdown of a task into its mode and mode parameter set b) Guided exploration by bounded permissible state space around the local neighborhood of the oracle's reference c) Mode encoder: an LSTM autoencoder trained on a custom modal dataset by minimizing reconstruction loss d) multi-mode policy trained with oracle guided policy optimization on a task environment 
 }
 
	\label{fig:afo}
	\vspace{-5mm}
\end{figure*}

This section presents the design methodology using OGMP for the bipedal control tasks: parkour and dive, as shown in Fig \ref{fig:afo}.  For a given task, we first define the task-vital modes—such as jump, leap, and pace for parkour—and design a reference generation scheme for each (Fig. \ref{fig:afo}.a).Spanning the mode parameter set, we employ the oracle to generate a custom dataset of diverse behaviors and train a mode encoder to construct a compact latent space for command conditioning (Fig \ref{fig:afo}.c). Finally, we train a multi-mode policy guided by the oracle (Fig \ref{fig:afo}.d). During policy optimization, the oracle is periodically queried to generate references online, bounding the policy's search space to a reference's local neighborhood for effective exploration(Fig \ref{fig:afo}.b) The above approach is explained in detail as follows. 

\subsection{Task description:} We apply the proposed framework to two bipedal control tasks: parkour and dive, with varying objectives and extent of multi-modality as shown in table \ref{table:task_def}. Note that the choice of task-vital multi-mode is user-defined, and table \ref{table:task_def} simply reflects our choice for the same. Evident from table \ref{table:task_def}, $J_\mathcal{T}$ is task-dependent. Recent attempts in quadruped parkour \cite{hoeller2024anymal,cheng2023extreme}, and locomotion \cite{miki2022learning,siekmann2021blind} show some well-shaped candidates for $J_\mathcal{T}$, albeit case-specific. In general, a reasonable $J_\mathcal{T}$ could be hard to design (for instance, the dive task); a compelling unified alternative would be to ``track" the oracle's $\epsilon$-neighbourhood reference to the optimal solution. Hence, we propose minimizing the task-independent surrogate objective: \( \Tilde{J_\mathcal{T}} := \sum_{t=0}^{\infty}\|x^{\pi}_t - x^{\Xi}_t\| \).
$\Tilde{J_\mathcal{T}}$'s applicability is studied in Sec. \ref{subsec:analysis}  and a reward is proposed in Sec \ref{subsec:mmode_policy} for an equivalent maximization objective.



        
                      


\begin{table}
    \centering
    \begin{tabular}{|c|c|c|}
        \hline 
        \multirow{2}{*}{$\mathcal{T}: J_\mathcal{T}$} & \multicolumn{2}{c|}{($m, \Psi_{m}) \in (\mathbb{M},\Psi_{\mathbb{M}}$)} \\
        \cline{2-3}
        & Mode & Parameters \\
        \hline
        \multirow{2}{*}{dive: $360 ^{\degree}$ flip and land} & settle & $\{\}$ \\
        & flip & $\{(r, h)\}$ \\
        \hline
        \multirow{3}{*}{parkour: traverse the track indefinitely} & pace & $\{v\}$ \\
        & jump & $\{(w, h)\}$ \\
        & leap & $\{(w, d)\}$ \\
        \hline
    \end{tabular}
    \vspace{2mm}
    \caption{Task description and corresponding task vital multi-modality. Parameters visualized in Fig. \ref{fig:orac_trajs}}
    \label{table:task_def}
    \vspace{-5mm}
\end{table}

 \begin{figure}
    \centering
        \centering

          \includegraphics[width=0.45\textwidth]{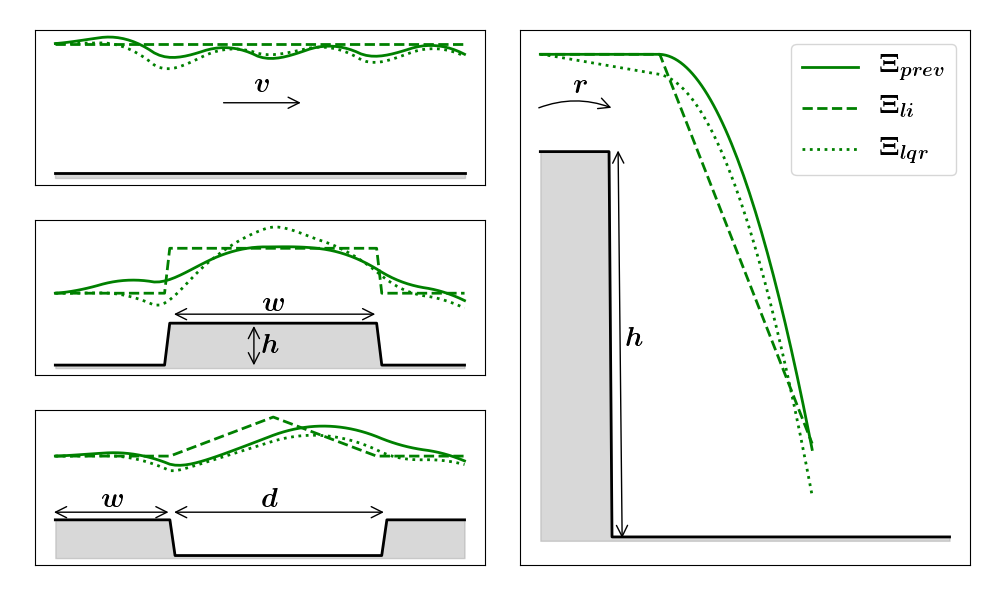}
        	\caption{Base height reference trajectories from various oracles for different modes}
        	\label{fig:orac_trajs}
         \vspace{-4mm}
 \end{figure}
\subsection{Oracle Design} For any locomotion task, a simple heuristic for an oracle would be linearly interpolating the relevant state variables from the initial to the desired goal states. In parkour, to advance along the track, the heading position can be linearly interpolated along the track while adapting to the terrain height, as shown in Fig. \ref{fig:orac_trajs} (left). For dive, the oracle can linearly interpolate the base height and corresponding rotational DoF from $0^{\degree}$ to $360^{\degree}$. 
Naming this heuristic oracle as $\Xi_\text{li}$ (Fig. \ref{fig:orac_trajs}), its high $\epsilon$ is obvious as the generated ansatz do not consider the system's inertia and gravity. Hence oracles that capture the dominant dynamics of the hybrid system are required. To this end, we have used a modified version of the simplified Single Rigid Body (SRB) model\cite{f&mmpc} whose dynamics in the world coordinates are given by: 
\par
{\footnotesize
\begin{align}
\label{eq:srb}
         m(\ddot{p}+g) = \sum_{i=1}^{2} f_i , \frac{d}{dt}(I\omega)\approx I\dot{\omega} = \sum_{i=1}^{2} (r_i \times f_i + \tau_i) = \sum_{1}^{2}\bar{\tau }_i \\
\label{eq:ssm}
    x_{t+1} = Ax_{t} + Bu_{t},\: y_{t+1} = Cx_{t}, \: u_t=[f_{1},f_{2},\Bar{\tau}_{1},\Bar{\tau}_{2}]^T
\end{align}
}

where $\ddot{p}$, $\omega$ is the robot COM acceleration and the angular velocity, $r_i,\,f_i,\,\tau_i$ are the position, force and moment vectors from the $i^{th}$ contact point and $m,\,I,\,g$ are the mass, moment of inertia and gravity. Typically, $r_i$ is from a predefined contact schedule, leading to time-varying dynamics. Since, by definition, an oracle need not provide realistic control, we define an auxiliary control $\bar{\tau }_i$, encompassing the overall moment, making the dynamics time-invariant. Additionally, the rotation and rotation rate matrices are made constant by considering the average reference orientation over a horizon. Upon these approximations to Eq. \ref{eq:srb} and discretization leads to a  linear time-invariant (LTI) system over the current horizon, where $x_t \in \mathbb{R}^{13},\,y_t,\, u_t\in\mathbb{R}^{12}$ are the gravity-augmented state, relevant output, and control vectors. Thus, oracles can be constructed considering two distinct phases: flight and contact. In flight, $u_t = 0$ as there are no contacts, and during contact, an optimizer of choice can be used to compute the optimal control for a given objective, $u_t = u^*_t$. The reference state trajectory is obtained by applying the corresponding control and forward simulating Eq. \ref{eq:ssm}. 
Using $\Xi_\text{li}$ as the reference for a quadratic tracking objective on the LTI system, optimizing with preview control \cite{cent_prev} and LQR results in oracles, $\Xi_\text{prev}$ and $\Xi_\text{lqr}$ respectively, having a smaller $\epsilon$ than $\Xi_\text{li}$ as shown in Fig \ref{fig:orac_trajs}

\begin{figure*}[h]
    \centering
    \begin{subfigure}[t]{\textwidth}
        \centering
        \includegraphics[width=\textwidth,height=0.05\textwidth]{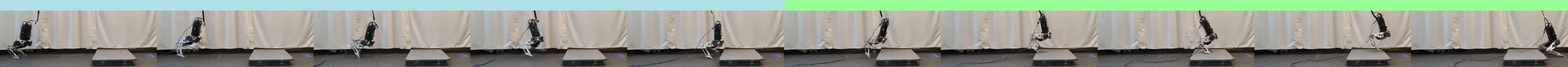}
        \caption{mode transitions: pace $\rightarrow$ jump}
        \label{fig:hw_p2j}
    \end{subfigure}
    \begin{subfigure}[t]{\textwidth}
        \centering
        \includegraphics[width=\textwidth,height=0.05\textwidth]{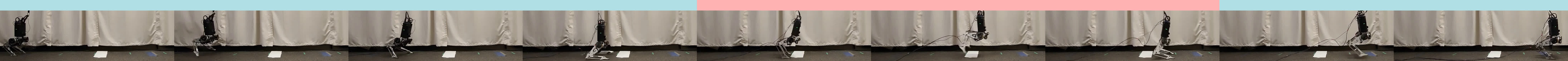}
        \caption{mode transitions: pace $\rightarrow$ leap  $\rightarrow$ pace}
        \label{fig:hw_p2l2p}
    \end{subfigure}
    \caption{Keyframes of an OGMP ($\pi_\text{parkour}$) demonstrating mode transitions (left to the right) with the color of the active mode marked: blue for pace, red for leap, and green for jump. }
    \label{fig:hw_train}
\end{figure*}

\subsection{Multi-mode Policy:}
\label{subsec:mmode_policy}
\textbf{Mode Encoder:} We train a mode encoder, $\xi$, on diverse locomotion modes to obtain a compact conditioning space ideal for commanding our policy. Similar to \cite{krishna2023learning}, the encoder, $ z = \xi(x[t,\,t+\Delta t])$, maps the trajectory space to a latent space ($\dim(z)=2$). Uniformly sampling from a mode parameter set,$\Psi_\mathbb{M}$ and a set of initial states, $\mathbb{X}_0$, we generate a rich and balanced modal dataset by querying the oracle, $\Xi$ as shown in Fig. \ref{fig:afo}.c. Minimizing the reconstruction loss for a $32$ neurons single hidden layer LSTM auto-encoder on the custom dataset generates a set of latent mode points with a structured clustering as visualized in Fig. \ref{fig:afo}.c. 

 \textbf{Mode Conditioned Policy:} Our choice of action space for a stationary policy is from \cite{krishna2023learning}. 
 Given the inherent partial observability of the system, 
 for the observation space $o_t \in \mathcal{O}$ we choose $o_t=[\Tilde{x}_t,z_t,\,c_t,h_t]$, where $\Tilde{x}_t$ is the robot's proprioceptive feedback, $z_t$ is the latent mode command, $c_t$ is a clock signal \cite{krishna2023learning} and $h_t$ is optional task-based feedback (like terrain scan for parkour). The per-step reward for a task-agnostic surrogate tracking objective is defined as 
\begin{equation}
\begin{aligned}
    r_t &:= r_\text{track} \text{ + } r_\text{regulation} \\
    r_\text{regulation} &:= 0.05 e^{-0.01\| u_t \|}   \text{ - }  0.3\cdot\mathds{1}\text{(if non-toe contact)}  \\
    r_\text{track} &:= 0.475 e^{-5\| er_p \|} \text{ + } 0.475 e^{-5\| er_o \|} 
\end{aligned}
\end{equation}
where $er_p, er_o$ are the errors in base position and orientations. 
Thus, $r_\text{track}$ minimizes an error in the task space, and $r_\text{regulation}$ regularizes the motion for enhancing sim-to-real. Note for the diverse modes across both the tasks (parkour and dive) the reward weights remain the same, hinting a sense of algorithmic robustness that arises from guided learning. The proposed permissible state constraint (Sec. \ref{sec:ogmp}) is programmed as a termination condition, terminate episode:= $\mathds{1}{( \|x^{\pi}_t - x^{\Xi}_t\|_W> \rho )}$. $W_{11}$ and  $W_{33}$ are set to $1$ for the parkour and dive tasks, respectively, with the remaining entries as zero. For solving the resulting random horizon POMDP, we use off-the-shelf PPO to train a policy: $128$ nodes per layer, $2$ layer LSTM network, where each episode is an arbitrary variant of the task $\psi_\mathcal{T}$ uniformly sampled from $\Psi_\mathcal{T}$.

%% file: sections/results.tex
\section{Results}
\label{sec:results}

\subsection{Performance}

To achieve extreme agility and mode versatility, a single multi-mode policy is trained per task: $\pi_\text{parkour}$ for parkour and $\pi_\text{dive}$ for diving. The supplementary video and Fig \ref{fig:ov} show $\pi_\text{parkour}$ successfully navigating challenging parkour tracks with blocks and gaps placed randomly, demonstrating versatile agility over leap lengths and jump heights. $\pi_\text{dive}$ performs omnidirectional flips from different heights and transitions smoothly to landing. Despite lacking a reference for the actuated DoFs, $\pi_\text{dive}$ learns an emergent behavior to curl and extend its legs for flips and landings, modulating the torso angular velocity and landing impact. As seen in the video, $\pi_\text{parkour}$ significantly deviates from the oracle's reference to find the optimal behavior, yet results in regularized motion due to the oracle bound. Sim-to-real transfer of  $\pi_\text{parkour}$'s modes and transitions can be seen in Fig \ref{fig:ov},\ref{fig:hw_train} and supplementary video. 

\subsubsection{Agility}For quantitative benchmarking of agility, we report the sample mean of performance metrics: Maximum Heading Acceleration (M.H.A), Froude number (M.F), Maximum Heading Speed (M.H.S), Episode Length (E.L), measured over $100$ episode rollouts in table \ref{table:metrics}. We define a test environment with a track length of $10$m, an episode length of $400$ steps for the parkour, and an episode length of $150$ for the dive. In each case, the episode is terminated only if the episode length is reached or the robot falls down (terrain-relative base height  $<0.3$ m).

\begin{table}[h]
  \centering
  \begin{tabular}{|c|c|c|c|c|c|c|}
    \hline


    metric, units  &
    $z_t$ & 
    $z_t,c_t$& 
    $h_t$ & 
    $h_t, c_t$ & 
    $h_t, z_t$ & 
    $h_t, z_t, c_t$ \\
    \hline


    M.H.A&
    \textbf{4.7g} &
    3.5g &
    3.2g &
    3.6g &
    3.5g &
    3.1g \\ 
    \hline
    M.H.S ($v$)& 
    1.4&
    1.57& 
    1.66& 
    1.74&
    1.74&
    \textbf{1.77} \\  
    \hline
    M.F ($\frac{v^2}{g \cdot ll}$) &
    0.48 & 
    0.56 & 
    0.64 & 
    0.69 &
    0.70 &
    \textbf{0.72} \\ 
    \hline

    \% E.L&
    0.18 &
    0.43 &
    0.63 &
    0.80 &
    0.66 &
    \textbf{0.84} \\
    \hline

  \end{tabular}
  \caption{Estimated metrics for variants of $\pi_\text{parkour}$ with different observation spaces}
\label{table:metrics}

  \vspace{-4mm}
\end{table}

On average, we find $\pi_\text{parkour}$ to reach accelerations of $4.7g$ with heading speeds of $1.77$ m/s, and Froude numbers between $0.48-0.72$  while completing $84\%$ of the track as shown in table \ref{table:metrics}. $\pi_\text{parkour}$ dynamically advances along the track, avoiding conservative motions with precise foot placements for landing and take-off, leading to agile manevours. The measured Fraude numbers and resulting motion are consistent with \cite{alexander84gaits}, where a switch from energy-efficient walking to agile jumping gaits was observed in bipedalism for a value around $0.5$. 

\subsubsection{Mode Versatility}
\label{subsec:mpst}

Since the defined $\Psi_\mathbb{M}$ of our training is a compact set, we leverage it to visualize the generalization of $\pi_\text{parkour}$ over mode parameters. Dilating $\Psi_\mathbb{M}$ and defining higher test ranges for each parameter, we test for both in-domain (ID) and out-of-domain (OD) generalization. 
Discretizing this test set, we evaluate $\pi_\text{parkour}$ and plot the undiscounted returns obtained for blocks and gaps with varying dimensions in Fig \ref{fig:mpt}. Different blocks and gaps require jumps and leaps of varying magnitudes, showcasing our policies' versatility. The training sets $\Psi_\text{jump} \text{ and } \Psi_\text{leap}$ are the regions within the boundary marked in black in each plot. Thus, $\pi_\text{parkour}$ shows consistent performance for variants within the black boundary (ID) while also extrapolating its skills by jumping and leaping over unseen terrain variants outside the black boundary (OD) as seen in the supplementary video and Fig \ref{fig:mpt}.

\subsection{Ablations and Analysis}
\label{subsec:analysis}
Finally, we analyze our choice of surrogate $\Tilde{J}_\mathcal{T}$ and design choices that impact performance (measured via undiscounted returns). For parkour, a potential true objective, $J_\mathcal{T}$, is the displacement along the track. Thus the validity of using $\Tilde{J}_\mathcal{T}$ can be quantified through its disparity with $J_\mathcal{T}$.

\begin{figure*}[h]
    \centering
    \begin{subfigure}[t]{0.6\textwidth}
        \centering
        \includegraphics[width=\textwidth]{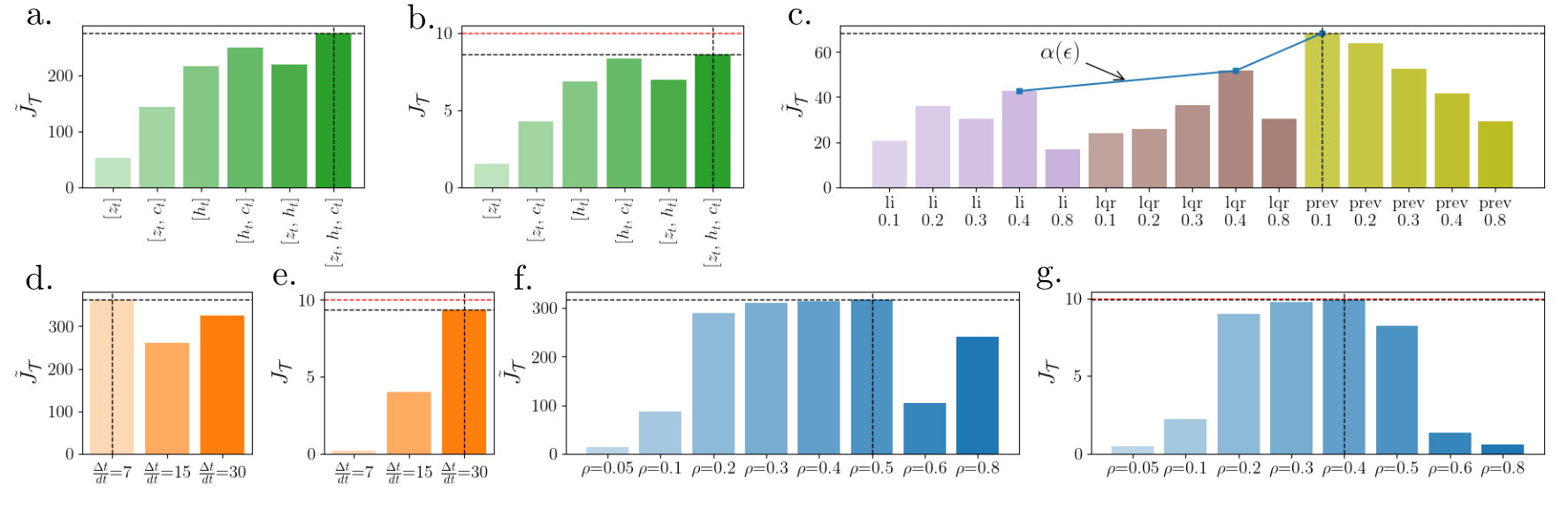}
        \caption{The estimated surrogate ($\tilde{J}_\mathcal{T}$) and true ($J_\mathcal{T}$) objectives vs. choice of 1) observation space, $o_t$ (a and b), 2) oracles with varied permissible state bound,  $\genfrac{}{}{0pt}{}{\Xi}{\rho,}$ (c), 3) horizon length, $\Delta t$ (d and e) and 4) permissible state bound, $\rho$ (f and g). In b, e, and g the upper bound of $J_\mathcal{T}$ is marked in red. }
        \label{fig:abl}
    \end{subfigure}
    \begin{subfigure}[t]{0.39\textwidth}
        \vspace{-3cm}
        \centering
        \includegraphics[width=\textwidth]{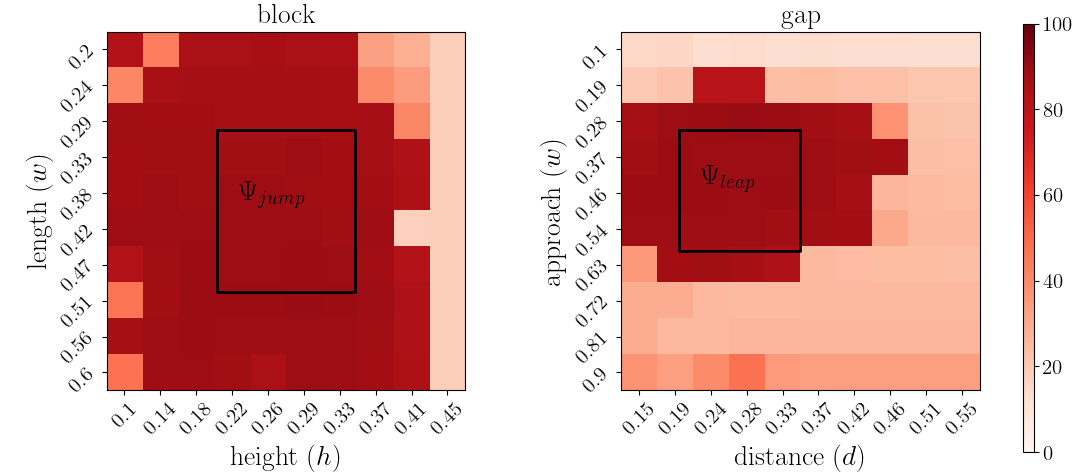}
        \caption{Heatmap of $\pi_\text{parkour}$'s undiscounted returns performing jumps and leaps of varying magnitudes both in and out of $\Psi_\mathbb{M}$ while traversing blocks and gaps of different dimensions}
        \label{fig:mpt}
    \end{subfigure}
    
    
    \begin{subfigure}[t]{0.6\textwidth}
        \centering
        \includegraphics[width=\textwidth]{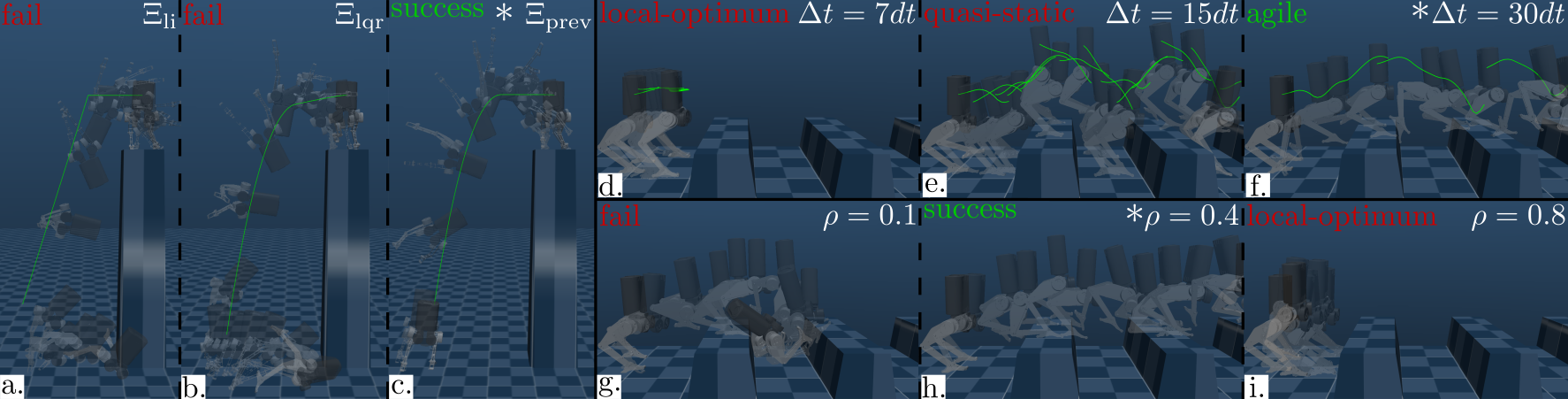}
        \caption{Motion traces comparing the choice of 1) oracles (a, b, c) where $\Xi_\text{li}$, $\Xi_\text{lqr}$  fail while only $\Xi_\text{prev}$ successful in a side flip dive from a $2$m block 2) horizon length (d, e, f) where the longest horizon ($\Delta t=30dt$) results in the desired agile performance 3) permissible state bound (g, h, i) where the tighter ($\rho=0.1$) and lenient ($\rho=0.8$) bounds fail}
        \label{fig:abl_trc}
    \end{subfigure}
    \begin{subfigure}[t]{0.39\textwidth}
        \centering
        \includegraphics[width=\textwidth,height=0.4\textwidth]{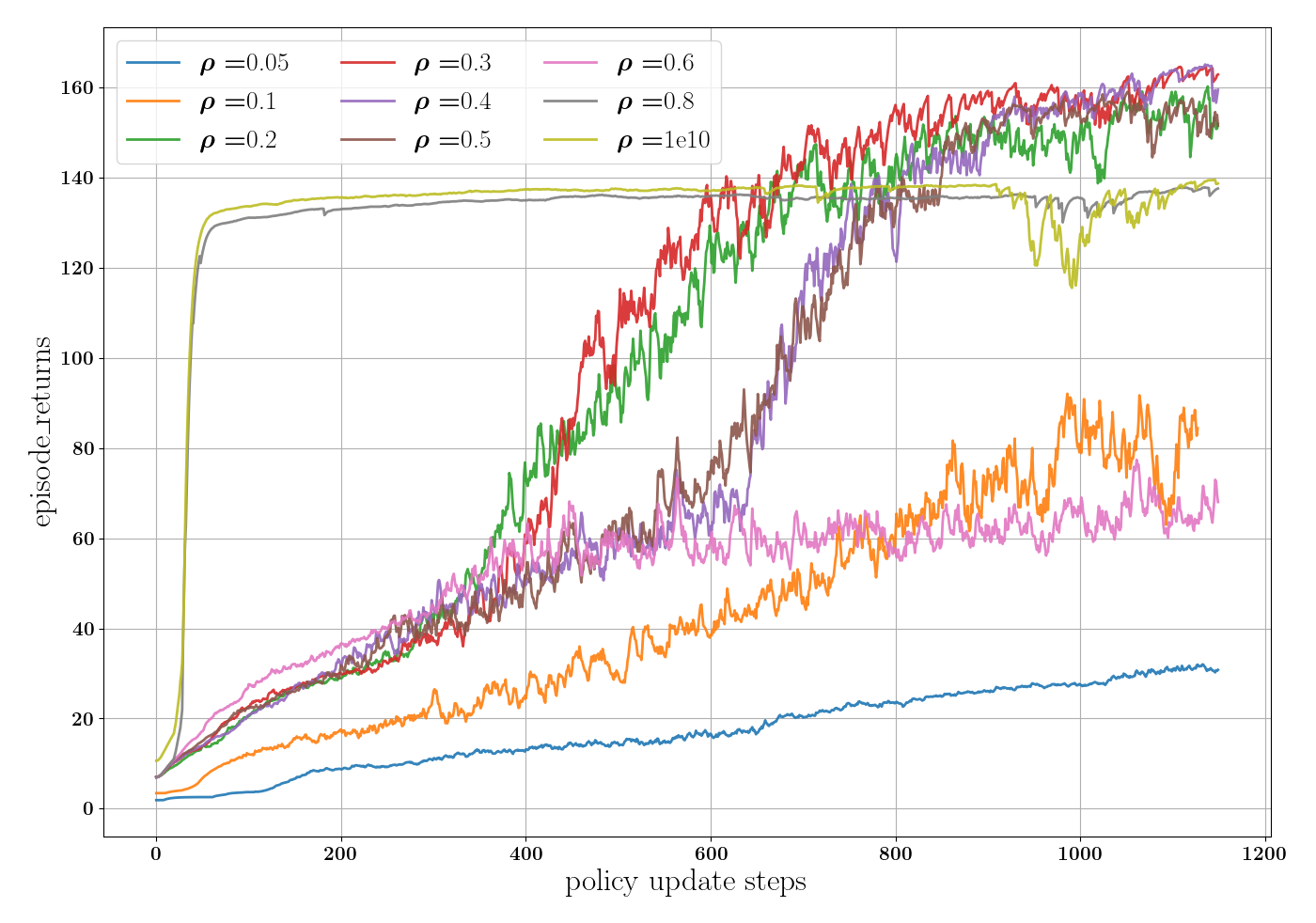}
        \caption{Training curves for different $\rho$'s with a fixed set of PPO's hyperparameters.}
        \label{fig:rho_vary}
    \end{subfigure}
    
    \caption{Abalations and Analysis}
    \label{fig:all_abl_analysis}
\end{figure*}

\textbf{Choice of observation space $(o_t)$}: Ablation of observation space components (excluding $\tilde{x}_t$) shows that variants with $c_t$ consistently outperform their counterparts (Fig. \ref{fig:abl}.a). Variants without $h_t$ are myopic and aggressive, with higher accelerations (table \ref{table:metrics})  as they purely rely on compressed $z_t$ for terrain feedback, making them sub-optimal compared to terrain-aware variants. From Fig. \ref{fig:abl}.a, we observe that latent conditioning does not improve performance (see [$h_t,c_t$] and [$z_t,h_t,c_t$]), hence is purely for analysis and reusability. However, a conditioned $\pi_\text{parkour}$ can use the oracle as a closed-loop reactive planner during inference, driving the system to the commanded mode. 
From Fig.\ref{fig:abl}.a and b, a similar trend of $J_\mathcal{T}$ and $\Tilde{J}_\mathcal{T}$ shows no disparity caused by observation space variations.


\textbf{Choice of oracle $(\Xi)$}: The three oracles, $\Xi_\text{li}$, $\Xi_\text{lqr}$, and $\Xi_\text{prev}$, have a non-increasing trend in maximum deviation bound, $\epsilon$. As the optimal exploration bound ($\rho^*$) depends on $\epsilon$, we vary $\rho$ from 0.1 to 0.8 for each oracle. For parkour, different oracles show no significant performance difference (see video results). However, for dive, $\Xi_\text{prev}$ variants perform significantly better (Fig. \ref{fig:abl}.c and \ref{fig:abl_trc}.c).A tighter bound ($\rho=0.1$) works best for $\Xi_\text{prev}$, as it has the lowest $\epsilon$. Conversely, higher exploratory deviation ($\rho=0.4$) performs best for $\Xi_\text{lqr}$ and $\Xi_\text{li}$. 
Thus, $\epsilon_\text{li} \geq \epsilon_\text{lqr} \geq \epsilon_\text{prev} \implies \rho^*_\text{li} \geq \rho^*_\text{lqr} > \rho^*_\text{prev}$, affirming Eq \ref{eq:rho_suff} .

\textbf{Choice of oracle's horizon $(\Delta t)$}: We evaluate policy performance across different horizons: $\Delta t = 7dt$, $15dt$, and $30dt$ for parkour. The shortest horizon, $\Delta t = 7dt$, leads to a myopic behavior, maintaining a high $\Tilde{J}_\mathcal{T}$ but low $J_\mathcal{T}$ as it remains stationary by exploiting the high replanning frequency (Fig. \ref{fig:abl_trc}.d and \ref{fig:abl}.d, e) as also observed by \cite{jenelten2023dtc}. Although advancing forward,  $\Delta t = 15dt$ fails to anticipate farther terrain, resulting in quasi-static maneuvers (Fig. \ref{fig:abl_trc}.e). Conversely, $\Delta t = 30dt$ enables the robot to leap efficiently from block to block, demonstrating agility and achieving the optimal outcome (Fig. \ref{fig:abl_trc}.f). Increasing $\Delta t$ aligns $\Tilde{J}_\mathcal{T}$ with the true task objective, $J_\mathcal{T}$ (Fig. \ref{fig:abl}.d, e).

\textbf{Choice of permissible state bound $(\rho)$}: For $\Xi_\text{prev}$ in parkour, we varied $\rho$ from $0.05$ to $0.8$ (Fig. \ref{fig:abl}.f and g). We found an optimal $\rho^* = 0.5$ with performance decreasing away from this value. For $\rho < \rho^*$, the optimal solution may lie outside the $\rho$-neighborhood of $x^\Xi$ (Fig. \ref{fig:abl_trc}.g and Fig. \ref{fig:rho_vary}). Conversely, $\rho > \rho^*$ increases local optima within $\epsilon+\rho$, leading to sub-optimal solutions (high $\Tilde{J}_\mathcal{T}$), seen in Fig. \ref{fig:abl_trc}.i for $\rho=0.8$, where the policy stagnates without advancing (low $J_\mathcal{T}$). Training curves in Fig. \ref{fig:rho_vary} show that $0.1 \leq \rho \leq 0.6$ converges to the global optima, while the rest settle at local optima. Note that $\rho \rightarrow \infty$ is standard PPO as it optimizes $\Tilde{J}_\mathcal{T}$ unguided by $\Xi$. Vanilla unguided PPO ($\rho = 10^{10}$) falls into the same local optima as $\rho = 0.8$. Thus, oracle-guided optimization improves standard PPO by escaping local optima with the right choice of $\rho$.


%% file: sections/conclusion.tex
\section{Conclusion } 
\label{sec:conclusion}
This paper introduces a framework for guided policy optimization through prior-bounded permissible states and task-vital multi-modality to tackle complex tasks. A single OGMP (per task) successfully solved agile bipedal parkour and diving, showcasing versatile agility. Future work will aim to extend to contact-rich open-world loco-manipulation tasks. Furthermore, by restricting the reachable states to a subset of the state space, we forgo the possibility of OGMP being a global policy. Consequently, any state outside the $\rho$-neighbourhood of the oracle's references may result in failure. Since current Deep RL methods lack any global convergence guarantees, this is nonrestrictive but highlights the need for future extensions for stronger algorithms.